# A Fuzzy expert system for goalkeeper quality recognition

Mohammad Bazmara[1], Shahram Jafari[2] and Fatemeh Pasand[3]

[1]School of Electrical and Computer Engineering, Shiraz university,
Shiraz,Iran

[2]School of Electrical and Computer Engineering, Shiraz university,
Shiraz,Iran

[3]School of Education and Psychology, Shiraz university,
Shiraz, Iran

**Abstract**

Goalkeeper (GK) is an expert in soccer and goalkeeping is a complete professional job. In fact, achieving success seems impossible without a reliable GK. His effect in successes and failures is more dominant than other players. The most visible mistakes in a game are those of goalkeeper's. In this paper the expert fuzzy system is used as a suitable tool to study the quality of a goalkeeper and compare it with others. Previously done researches are used to find the goalkeepers' indexes in soccer. Soccer experts have found that a successful GK should have some qualifications. A new pattern is offered here which is called "Soccer goalkeeper quality recognition using fuzzy expert systems". This pattern has some important capabilities. Firstly, among some goalkeepers the one with the best quality for the main team arrange can be chosen. Secondly, the need to expert coaches for choosing a GK using their senses and experiences decreases a lot. Thirdly, in the survey of a GK, quantitative criteria can be included, and finally this pattern is simple and easy to understand.

*Keywords*: soccer goalkeeper, fuzzy logic, indexes, quantitative criteria, expert system.

## 1. Introduction

Most parents are interested to see their children successful in a sport major, for instance, as a national or international champion. This interest is mostly towards soccer, the world's most popular sport. In soccer, which is a group sport, GK has a key role. GK is an expert and goalkeeping is a complete professional job. This doesn't exclude GK from other players, in fact it shows the undeniable effect of GK and achieving success seems impossible without having a reliable one. His effect in successes and failures is more dominant than other players. The process of discovering qualified goalkeepers to attend organized practice programs is one of the most important things which have appeared in soccer for years.

In this paper, first of all, goalkeeping in soccer will be discussed and then the scientific and experimental indexes and criteria of GKs in soccer are evaluated and their application and effectiveness in practice will be surveyed. At last, it's relation to expert fuzzy system and use of these rules in finding the quality and performance of GKs is offered.

In general, there is no scientific and applied equation for recognition of GK performance and efficiency in soccer. This procedure is done by coaches using their experiences and observations of players. This process turns out to be a problem when for choosing the main GK of a national team, there is some GKs in the same level. As you know there are many choices for a national team GK and different coaches might choose differently. The existence of GKs in the same level makes this selection process difficult. Like the selection of Buffon or Toldo in Italy, 8 years ago, Oliver Kan or Leman in Germany, 8 years ago, Abbiati or Dida in Milan FC, 5 years ago.

In an expert system, fuzzy rules are used to form a pattern for surveying the goodness and quality level of GKs. Membership functions are in the center of these fuzzy patterns, these functions and rules are formed and offered on the basis of experts' knowledge. An expert is a person whose knowledge in a special field is gained gradually and by learning and experiencing. The aim in this paper is designing an expert fuzzy system on the basis of experts' knowledge in order to eliminate wrong recognitions based on experience in choosing the better GK and increasing the efficiency in this process. As you know, this subject has never been surveyed in literature.

## 2. Using fuzzy rules for GK quality recognition

The use of fuzzy systems has had the most growth in systems engineering [1]. To some extents, fuzzy means apposite to exact. Some concepts like GK's quality which can't be defined accurately or don't have clear boundary in time and place, are considered as fuzzy concepts. On the other hand, fuzzy logic is pertained to the definition of fuzzy sets [2]. In this method, the membership of a member in a fuzzy set is measured by the membership function average, where the values are between 1 (complete membership) and 0 (no membership).

Besides that, Prof Lotfi Zadeh has stated in the Principle of "incompatibility" that, as the complexity of a system increases, the person's ability to make precise and yet relevant statements about his behavior diminishes until a threshold is reached beyond which precision and relevance becomes almost mutually exclusive characteristics [3]. Now it can be understood that, real complex problems need smart systems which combine knowledge, techniques and methods from different references [4].

Ecologic surveys are known naturally complex [5]. Therefore it seems that fuzzy values are suitable and persistent techniques for solving the duality (0, 1) in variations of quality and efficiency [6].Fuzzy sets theory offers a more realistic representation of correct imaginations (compared to the two methods of fuzzy rules and values with algebraic and bullion equations). This main factor in transition from accurate sets to fuzzy sets is membership function [7]. Membership functions determine the integer values for statements like "a player is a bit more than average better for goalkeeping "or more complex statements that are used in daily life.

Because of the complex and vague essence of defining the GK quality, fuzzy logic can be useful in evaluation of these cases:
- Selection of important GK indexes.
- Surveying the values and importance of the above indexes.
- Decision making by decision makers and coaches.

## 3. fundamentals of fuzzy set and operators

The mathematics of fuzzy sets and fuzzy logic is discussed in detail in many books [7,8,9]. Here, we only discuss certain basic aspects concerning the mathematics that underlay fuzzy logic. We try to provide the minimal information needed to understand the construction method and the general working of the fuzzy model introduced later on.

### 3.1. From crisp to fuzzy sets

Let $U$ be a collection of objects u which can be discrete or continuous. $U$ is called the universe of discourse and $u$ represents an element of $U$. A classical (crisp) subset $C$ in a universe $U$ can be denoted in several ways like, in the discrete case, by enumeration of its elements: $C = \{u_1, u_2,..., u_p\}$ with $\forall i: u_i \epsilon U$. Another way to denote $C$ (both in the discrete and the continuous case) is by using the characteristic function $X_F: U \rightarrow \{0, 1\}$ according to $X_F(u) = 1$ if $u \in C$, and $X_F(u) = 0$ if $u \notin C$. The latter type of definition can be generalized in order to define fuzzy sets. A fuzzy set $F$ in a universe of discourse $U$ is characterized by a membership function $\mu_F$ which takes values in the interval [0, 1] namely, $\mu_F: U \rightarrow [0,1]$.

### 3.2. Operators on fuzzy sets

Let $A$ and $B$ be two fuzzy sets in $U$ with membership functions $\mu_A$ and $\mu_B$, respectively. The fuzzy set resulting from operations of union, intersection, etc. of fuzzy sets are defined using their membership functions. Generally, several choices are possible:

*Union:* The membership function $\mu_{A \cup B}$ of the union $A \cup B$ can be defined by $\forall u: \mu_{A \cup B} = max\{\mu_A(u), \mu_B(u)\}$ or by $\forall u: \mu_{A \cup B} = \mu_A(u) + \mu_B(u) - \mu_A(u)\mu_B(u)$.

*Intersection:* The membership function $\mu_{A \cap B}$ of the union for all $A \cap B$ can be defined by $\forall u: \mu_{A \cap B} = min\{\mu_A(u), \mu_B(u)\}$ or by $\forall u: \mu_{A \cap B} = \mu_A(u)\mu_B(u)$

*Complement:* The membership function of the complementary fuzzy set $A^c$ of $A$ is usually defined by $\forall u: \mu_{A^c} = 1 - \mu_A(u)$.

### 3.3. Linguistic variables

Fuzzy logic enables the modeling of expert knowledge. The key notion to do so is that of a linguistic variable (instead of a quantitative variable) which takes linguistic values (instead of numerical ones). For example, if the height is a linguistic variable, then its linguistic values could be one from the so-called termset T(height) = {short, tall} where each term in T(height) is characterized by a fuzzy set in the universe of discourse, here, e.g., $U$ = [0, 10].

Therefore, these linguistic values are characterized by fuzzy sets described by a membership function as shown in Figure 1.

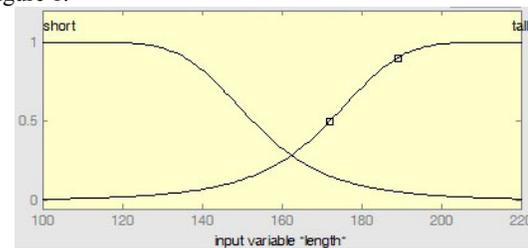

Fig. 1: Schematic representation of linguistic values "short" and "tall" by the corresponding membership functions for the linguistic variable "height rate" of a GK.

### 3.4. Knowledge representation by fuzzy IF-THEN rules

Fuzzy logic enables the formulation of prototypical linguistic rules of a fuzzy model that can easily be understood by experts where, at the same time, all kinds of mathematical details are hidden. To do so, knowledge is represented by fuzzy IF-THEN linguistic rules having the general form:

If $x_1$ is $A_1$ AND $x_2$ is $A_2$ ... AND $x_m$ is $A_m$ THEN y is B
AND $x_m$ is $A_m$ THEN y is B; where $x_1...x_m$ are linguistic input variables with linguistic values $A_1, ..., A_m$, respectively and where $y$ is the linguistic output variable with linguistic value $B$.

To illuminate we consider animal units and plantation density as the principal factors for having equilibrium. Then the relevant fuzzy rules could be:

- IF height of GK is tall AND Flexibility is bad THEN GK is Level3,
- IF height of GK is short AND Flexibility is good THEN GK is Level6.

### 3.5. Architecture of fuzzy systems

Fuzzy inference systems or, shortly, fuzzy systems (FSs) usually implement a crisp input–output (I–O) mapping (actually, a smooth function O = f (I)) consisting of basically four units, namely:
- a Fuzzifier transforming crisp inputs into the fuzzy domain,
- a rule base of fuzzy IF-THEN rules,
- an inference engine implementing fuzzy reasoning by combining the fuzzified input with the rules of the rule base,
- A Defuzzifier transforming the fuzzy output of the inference engine to a crisp value (Figure 2).

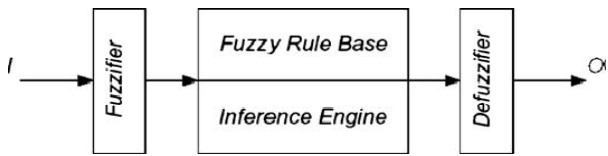

Fig. 2: Building blocks of a Fuzzy Inference System.

In some practical systems, the Fuzzifier or the Defuzzifier may be absent.

### 3.6. Fuzzy reasoning

Probably the hardest part to understand is the precise way fuzzy reasoning can be implemented. An extensive discussion of this topic is outside the scope of this paper so we limit ourselves here to present just the basic idea. Classical logic is our starting point using the classical reasoning pattern 'modus ponens':
Given fact ''x is A'' and rule ''IF x is A, THEN y is B'', we conclude that ''y is B''.
Applying fuzzy reasoning, classical modus ponens can be generalized to an 'approximate reasoning' scheme of type:
Given fact ''x is A' '' and rule ''IF x is A, THEN y is B'', we conclude that ''y is B' ''.
Here, the assumption made is that the closer $A'$ to $A$, the closer will $B'$ be to $B$. It turns out that especial combinations of operations on fuzzy sets like 'max–min' and 'max-product' composition can fulfill this requirement. The complete fuzzy reasoning in a FS can be set up as follows:
1. the fuzzification module calculates the so-called 'firing rate' (or degree of fulfillment) of each rule by taking into account the similarity between the actual input $A'$ defined by membership function $\mu_{A'}(x)$ and in case of a crisp input $X_p$ defined by the value $\mu_{A'}(X_p)$ and the input $A$ of each rule defined by membership function $A'(x)$.
2. Using the firing-rates calculation, the inference engine determines the fuzzy output $B'$ for each rule, defined by membership function $\mu_{B'}(y)$.
3. The inference engine combines all fuzzy outputs $B'$ into one overall fuzzy output defined by membership function $\mu(y)$.
4. The defuzzification module calculates the crisp output $y_p$ using a defuzzification operation like 'centroid of gravity (area)'.

For a treatment in depth on FSs, its construction and corresponding reasoning schemes (including the most popular systems like Mamdani[10] and Tagaki-Sugeno fuzzy models)[11]. we refer to the above-mentioned textbooks.

## 4. Research method

According to the researches done by the authors, GK has the main and most important post in soccer and without a reliable one, achieving success seems impossible. His effect in successes and failures is more dominant than others. The most visible mistakes in a game can be called to be those of the GKs'. So they deserve special attention. There are various definitions of a GK, each one considering a special viewpoint. Allen Hodkinson, Scottish national team GK coach, believes that a good GK may make an egregious mistake in every 6 matches, a very good GK may make one in every 9 matches and an excellent GK may make one in every 12 matches, while a superstar GK may make that mistake in every 15 matches.

Bruce Grobbelaar, the ex-GK of Liverpool and Zimbabwe national team says: "if you want to be a successful GK, first you should be a good gymnast, then practice basketball and handball, if you have the time practice cricket and baseball, and then do goalkeeping exercises.

Jean-Paul Sartre, French well known philosopher has some views about goalkeeping. He says: "goalkeeping is not a complex of acrobatic and conscious caprioles, but it is science, the science of determining angles, and practicing the angle to the ball".

As we can see, the definitions about GKs are so various and each follows a special viewpoint. Now, considering a GK's significant characteristics, an expert system with fuzzy rules is proposed here, that can be generalized for all GKs and teams. As you know, the GK's significant characteristics are as follows [12]:
1) Exit from the goal
2) Flexibility
3) The ability to repulse overhead shoots
4) Establishing connection
5) Courage
6) Effectiveness (leadership)
7) Success in person to person battles with rival's invaders.
8) Being tall.

These items' effectiveness and priority is never discussed in researches. We can suppose each item's effectiveness equal to the other one and design our fuzzy system on this basis.
.

## 5. Constructing the Fuzzy model

As mentioned previously there are 8 significant characteristics of a GK that can be considered as input parameters. Now we should assign fuzzy sets or linguistic variables to our inputs.

Table 1: input linguistic variables

| Input (characteristic) | Linguistic variables |
|---|---|
| Exit from the goal | Good-bad |
| Flexibility | Good-bad |
| Overhead dominance | Good-bad |
| Establishing connection | Good-bad |
| Courage | Good-bad |
| Leadership | Good-bad |
| person to person battles | Good-bad |
| Being tall | Tall-short |

We applied two linguistic variables to the inputs in order to be able to extract rules with high accuracy and independent to an expert. As you know, the less the linguistic variables are, the more the interpretability of the system will be, which makes it easy to understand. While extracting the characteristics, no data was found to show that they have any priority or recency or different significance towards each other. So they can be said to have equal effectiveness for defining the GK quality.

Considering the 8 characteristics (which leads to a total of 250 rules), we can have their combinations to achieve outputs. If we assign 9 fuzzy sets or linguistic variables to the output or the GK quality, we can make use of statistical combinations to achieve rules. (See figure 3)

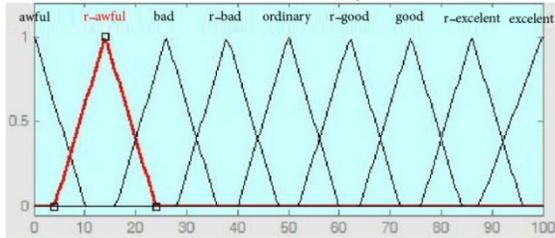

Fig. 3: final output linguistic variables

Using relational calculations we can see that there will be specific number of outputs for each linguistic variable. (See table 2)

In figure 4, the linguistic variables assigned to inputs except being tall are shown in the [0, 1] interval.

Table 2: different situations of rules

| Linguistic variable | No. of rules | Rule sample |
|---|---|---|
| Excellent | 1 | good-good-good-good-good-good-good-tall |
| Almost excellent | 8 | good-good-good-good-good-good-bad-tall |
| Good | 28 | good-good-bad-good-good-good-bad-tall |
| Relatively good | 56 | good-good-bad-good-good-good-bad-short |
| Ordinary | 70 | good-bad-bad-good-good-good-bad-short |
| Relatively bad | 56 | bad-good-bad-good-good-bad-bad-short |
| Bad | 28 | bad-bad-bad-good-good-bad-bad-short |
| Relatively awful | 8 | bad-bad-bad-bad-good-bad-bad-short |
| Awful | 1 | bad-bad-bad-bad-bad-bad-bad-short |

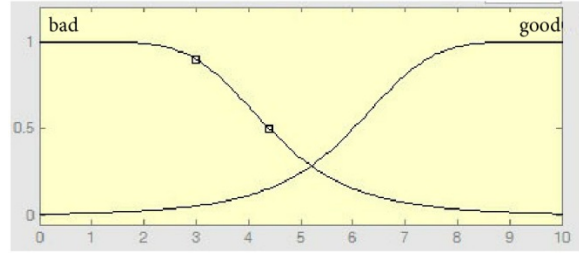

Fig. 4: input linguistic variables

As you can see in this figure, being tall is set into [100,220] cm interval. The users who are mostly soccer coaches can evaluate the GK quality by entering desired input. For each of the first 7 parameters they can input a value between 0 and 10, or use the expressions "good" or "bad", then for being tall they can input the player's height or use the linguistic expressions "tall" or "short". Finally the designed fuzzy model displays the final output which is the GK's quality level.

For instance, considering table 3, we want to compare 3 GKs, who have gained these values for the main mentioned characteristics, and the system has determined their quality.

Table 3: determination of 3 GK qualities using fuzzy model

| Characteristics | 1st GK | 2nd GK | 3rd GK |
|---|---|---|---|
| Exit from the goal | 7 | 6 | 6 |
| Flexibility | 4 | 7 | 5 |
| Overhead dominance | 7 | 5 | 7 |
| Establishing connection | 8 | 8 | 9 |
| Courage | 7 | 8 | 7 |
| Leadership | 9 | 9 | 9 |
| person to person battles | 4 | 3 | 6 |
| Being tall | 187 | 198 | 195 |
| Total sum | 66.1 | 67.9 | 70.7 |

Using this system, with no personal view and no use of personal experience, the 3rd GK can be chosen among 3 almost similar GKs as the best one. The calculated fuzzy model for the 3rd GK is shown in figure 5.

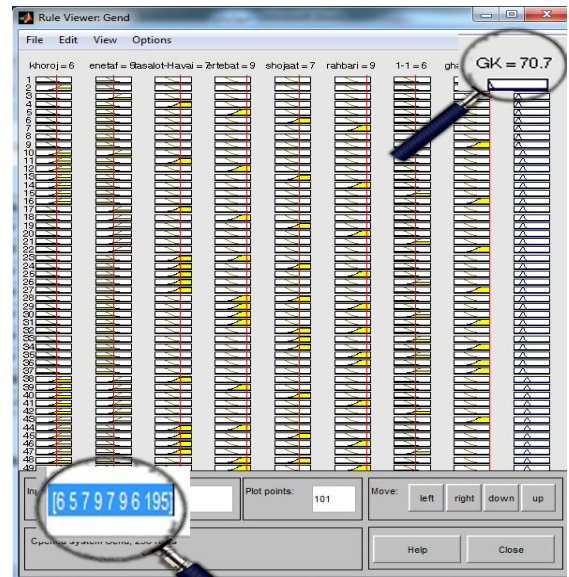

Fig. 5: final output display for the 3rd GK in Matlab

## 6. Conclusions

Considering the fact that soccer coaches use their personal experiences and viewpoints to choose suitable GKs, and no specific method or formula exists for choosing a qualified GK. It is possible that different coaches come up with different choices. Here, we proposed the fuzzy pattern that functions on the basis of significant characteristics of GKs and assigns a value between 0 and 100 to them. This method is much more efficient than classical methods based on personal experience and sense.